\documentclass[runningheads]{llncs}
\usepackage[T1]{fontenc}
\usepackage{amsmath,amssymb}
\usepackage{xcolor}
\usepackage{booktabs}
\usepackage{multicol}
\usepackage{multirow} 
\usepackage{hyperref}
\usepackage{graphicx,verbatim}
\begin{document}
\title{Thinking with Gaze: Sequential Eye-Tracking as Visual Reasoning Supervision for Medical VLMs}
\titlerunning{Thinking with Gaze}
%

\author{
    Yiwei Li$^{1}$, Yifan Zhou$^{1}$, Huaqin Zhao$^{1}$,Zihao Wu$^{1}$,  Zhengliang Liu$^{1}$, Xiang Li$^{3}$, Quanzheng Li$^{3}$, Tianming Liu and Lin Zhao$^{4,\dagger}$
    %
}  
\authorrunning{Yiwei Li et al.}
\institute{1. School of Computing, University of Georgia 2. Department of Computer Science and Engineering, University of Texas, Arlington 3. Massachusetts General Hospital, Harvard Medical School 4. Department of Biomedical Engineering, New Jersey Institute of Technology  \\
    \email{Corresponding author: Lin Zhao (email: lin.zhao.1@njit.edu).}}
  
\maketitle              
\begin{abstract}

Vision--language models (VLMs) process images as visual tokens, yet their intermediate reasoning is often carried out in text, which can be suboptimal for visually grounded radiology tasks. Radiologists instead diagnose via sequential visual search; eye-tracking captures this process as time-ordered gaze trajectories that reveal how evidence is acquired over time. We use eye-gaze as supervision to guide VLM reasoning by introducing a small set of dedicated gaze tokens. These tokens are trained to predict gaze-selected image patch indices in temporal order, encouraging the model to follow human-like evidence acquisition and integration. Experiments on MIMIC-EYE and multiple external zero-shot benchmarks show consistent gains over baselines, achieving state-of-the-art in-domain performance and improved out-of-domain robustness. These results highlight temporally ordered gaze as an effective supervision signal for learning visually grounded medical reasoning.


\end{abstract}

\section{Introduction}

VLMs have recently achieved remarkable progress by representing images as sequences of visual tokens and performing unified autoregressive reasoning in the token space~\cite{bai2025qwen3}. Despite multimodal inputs, however, many VLM pipelines still rely on text-only intermediate reasoning~\cite{liu2025visual}: the model first converts visual evidence into textual descriptions and then ``thinks'' in language~\cite{su2025thinking}. This can be suboptimal for tasks that are inherently visual, where the most informative signals are not easily verbalized without loss~\cite{liu2023interngpt}. A growing line of work therefore advocates thinking with images: either (i) augmenting inference with visual tools such as zooming, cropping, or drawing auxiliary cues, or (ii) internalizing visual reasoning by generating and manipulating latent visual tokens---a atent visual chain-of-thought (latent visual CoT) that remains grounded in visual representations rather than purely textual explanations~\cite{wang2026render}.

Medical imaging is a particularly compelling domain for such visually grounded reasoning~\cite{li2023llava}. Radiologists do not diagnose by reading off a static list of findings; instead, they follow a sequential visual search process, revisiting suspicious regions and integrating evidence over time~\cite{alayrac2022flamingo}. Crucially, eye-gaze naturally encodes this process as an ordered trajectory~\cite{karargyris2021creation}, providing a temporally structured supervision signal that is intrinsically closer to ``reasoning'' than a single global label. In other words, gaze is not merely an attention map~\cite{wang2024gazegnn}; it is a time-ordered record of how experts gather evidence, which is well aligned with the token-by-token computation of modern VLMs. This observation motivates us to treat gaze as a form of visual reasoning supervision and to inject it into VLM training in a continuous feature level.

In this work, we use eye-gaze as supervision to explicitly teach a pretrained VLM how radiologists gather evidence while interpreting chest X-rays. Using MIMIC-EYE~\cite{hsieh2023mimic}, we augment a fixed-format generation interface with a small set of dedicated gaze tokens that serve as intermediate evidence carriers. These tokens are supervised to predict gaze-derived patch indices in the same temporal order as the radiologist’s scanpath, so the model is encouraged to follow a human-like evidence acquisition routine—where to look next and how to accumulate findings—rather than relying on a static spatial attention prior. Empirically, this gaze-guided training consistently improves over strong instruction-tuned baselines, achieving state-of-the-art performance on MIMIC-EYE and stronger robustness on external, out-of-distribution benchmarks.

Our contributions are threefold:
\begin{itemize}
    \item \textbf{Gaze-guided reasoning supervision for radiology VLMs.} We introduce a lightweight mechanism that uses temporally ordered eye-gaze to supervise a small set of dedicated tokens, explicitly training the model to mimic radiologists' step-by-step evidence gathering and reasoning.
    \item \textbf{State-of-the-art accuracy with clinician-friendly interpretability.} Our approach achieves the best performance on MIMIC-EYE and delivers consistent gains over strong instruction-tuned baselines while producing gaze-linked patch evidence that supports case-level auditing and retrospective review.
    \item \textbf{Stronger out-of-domain robustness.} By learning human-like evidence acquisition patterns rather than dataset-specific shortcuts, the resulting VLM generalizes better to external benchmarks, demonstrating improved zero-shot transfer under distribution shift.
\end{itemize}

\begin{figure}
\includegraphics[width=\textwidth]{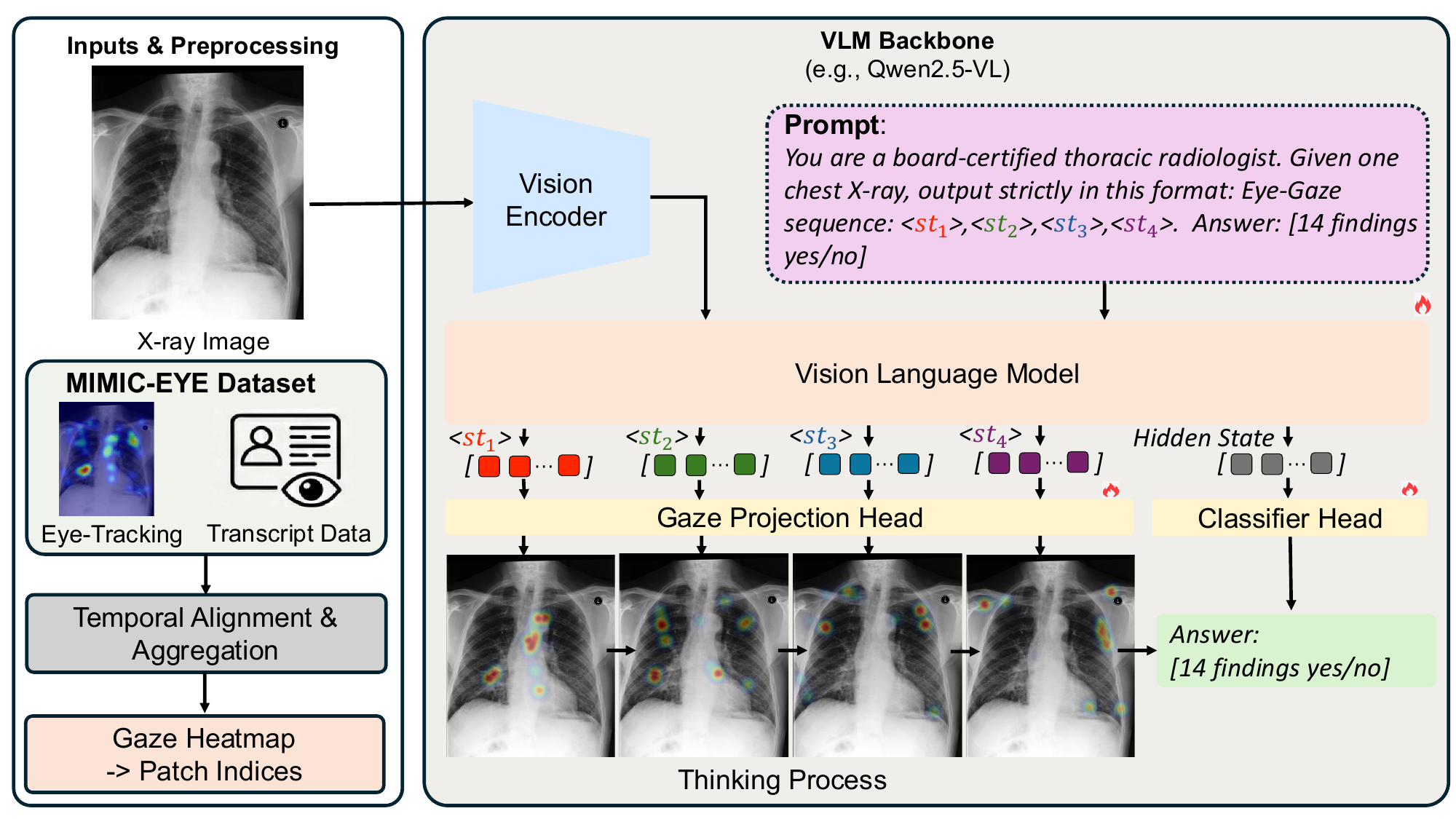}
\caption{\textbf{Method overview.} We fine-tune a pretrained VLM with MIMIC-EYE by injecting gaze supervision as discrete patch indices. Stage~1 learns four dedicated gaze tokens via a lightweight projection head that predicts gaze-selected patch IDs (cross-entropy). Stage~2 adds a 14-label classifier head to predict radiographic findings (binary cross-entropy) while enforcing a strict fixed-format yes/no output.}
\label{fig:method}
\vspace{-2mm}
\end{figure}

\section{Related Work}


\subsection{Thinking with Images and Latent Visual Reasoning}
Recent VLM research has begun to move beyond purely text-based chains-of-thought by explicitly incorporating visual manipulation into the reasoning process, often summarized as ``think with images''~\cite{openai_thinking_with_images}. Although VLMs take images as input, their intermediate reasoning is typically carried out in text~\cite{yang2023mmreact}, which can be suboptimal for visually grounded tasks. A pragmatic approach is tool-augmented visual reasoning, where the model invokes visual tools (e.g., zooming, cropping, rotating, drawing auxiliary lines, or sketching) during inference so that iterative visual operations and the resulting images become part of the reasoning workflow~\cite{wu2023visualchatgpt}. Such schemes can be categorized by how tool usage is determined and executed: (i) fixed workflows with predefined tools and order; (ii) model-driven tool selection (deciding when/what to call); (iii) fixed tool code with model-predicted parameters; and (iv) model-generated tool code enabling a broader space of visual operations~\cite{suris2023vipergpt}.

Beyond external tool calls, a complementary direction aims to internalize visual reasoning by operating in a latent visual space, often referred to as latent visual chain-of-thought. A representative example is CoVT~\cite{qin2025covt}, which argues that many VLMs suffer from a discrete text-space bottleneck: although images are provided as inputs, intermediate reasoning is largely confined to language tokens, potentially losing fine-grained perceptual cues. To mitigate this, CoVT enables reasoning in a continuous visual token space by generating and refining visual tokens as intermediate states, thereby keeping computation visually grounded rather than fully translating evidence into text. Training typically introduces auxiliary objectives (e.g., visual reconstruction or distillation) to ensure that the generated visual tokens remain faithful to the input image and preserve relevant visual evidence; such supervision can be obtained from auxiliary models, auxiliary images, or regions-of-interest (ROIs) derived from the original image (e.g., bounding boxes). Visual-token representations vary across methods, including ViT features (or projected features)~\cite{dosovitskiy2020image}, intermediate activations within the VLM, or discrete codes from VQ-VAE style tokenizers~\cite{oord2017vqvae}.

\subsection{Eye-gaze Supervision in Radiology}
Eye-tracking provides an objective signal of radiologists' visual search behavior~\cite{kundel1975interpreting} and has been studied for decades to characterize expert vs.\ novice reading patterns and error modes~\cite{kundel1978visual}.
Recently, gaze has also been incorporated into deep learning pipelines as an auxiliary supervision signal~\cite{samuel1995mechanism}, where gaze-derived attention can guide models toward diagnostically relevant regions~\cite{ganesan2018review} (e.g., gaze-guided lesion detection with CNNs, gaze pattern analysis in mammography).
However, systematically integrating gaze into vision--language alignment remains less explored~\cite{bhattacharya2022radiotransformer}: the key challenge is to transform temporally synchronized gaze and speech/transcripts into a supervision signal that improves cross-modal correspondence rather than only spatial localization~\cite{stember2020integrating}.

\vspace{-2mm}
\section{Method}
\subsection{Problem Formulation}
Given a chest X-ray image $I$ from MIMIC-EYE~\cite{hsieh2023mimic} and its associated gaze-derived supervision, our goal is to train a VLM to (i) output a structured radiology-style 14-label report in a strictly constrained format, and (ii) predict a 14-dimensional multi-label vector $\mathbf{y}\in\{0,1\}^{14}$ indicating the presence of common findings.
In addition, we leverage radiologists' eye-gaze signals to supervise where the model should attend, by aligning a small set of latent ``gaze tokens'' with image patch indices.

\subsection{Dataset and Multi-modal Preprocessing (MIMIC-EYE)}
We follow the multi-modal processing paradigm used in EGMA~\cite{ma2024eye} and use MIMIC-EYE as the training source.
MIMIC-EYE provides chest radiographs with synchronized eye-tracking signals and transcribed speech produced during diagnostic reading. We describe MIMIC-EYE as consisting of images extracted from MIMIC with each sample accompanied by eye-tracking data and transcript text, where eye-tracking is derived from the publicly available EYE GAZE and REFLACX~\cite{bigolin2022reflacx} datasets and modalities are synchronized in time.

\paragraph{Audio--text--gaze temporal alignment.}
Since the modalities are time-synchronized, the audio stream is temporally aligned with eye-gaze. We align transcript tokens by segmenting the audio into per-word time intervals (before/after each word’s pronunciation), enabling time-resolved text--gaze alignment. In practice, word-level pairing can be unreliable due to rapid speech (no gaze samples within a word interval) and intermittent gaze dropouts (e.g., blinks or head motion). To mitigate missing word-level signals while preserving semantics, we aggregate gaze at the sentence level when constructing gaze supervision.

\paragraph{From gaze to patch indices.}
We convert the gaze signal into attention heatmaps over the image to represent regions the radiologist focuses on, then discretize the heatmap into a fixed patch grid consistent with the VLM visual tokenizer.
We store gaze supervision as a set of patch indices (top-$k$ patches per gaze token), which is compact and directly compatible with a classification head over patch IDs.

\subsection{Model Architecture}
\paragraph{Backbone VLM.}
We use a pretrained VLM backbone (Qwen2.5-VL-7B-Instruct in our implementation~\cite{wu2025qwen}) consisting of a vision encoder and an autoregressive language model. Model overview can be seen in Figure~\ref{fig:method}.

\paragraph{Fixed-format generation with latent gaze tokens.}
To enable stable extraction of gaze-related hidden representations, we reserve exactly four special placeholder tokens (denoted as $\langle st\rangle_1,\dots,\langle st\rangle_4$) at the beginning of the assistant response.
The assistant is trained to output:
\begin{equation}
\langle st\rangle_1\langle st\rangle_2\langle st\rangle_3\langle st\rangle_4\ \texttt{Answer:}\ \text{14 findings as yes/no}.
\end{equation}
These four token positions act as latent ``gaze tokens'' whose hidden states are supervised to match gaze-selected image patches.

\paragraph{Gaze projection head.}
Let $\mathbf{H}\in\mathbb{R}^{T\times d}$ be the final-layer hidden states of the language model for an input sequence of length $T$.
We extract the hidden vectors at the four gaze-token positions:
\begin{equation}
\mathbf{z}_i = \mathbf{H}[p_i]\in\mathbb{R}^{d},\quad i\in\{1,2,3,4\},
\end{equation}
where $p_i$ denotes the token index of $\langle st\rangle_i$ in the concatenated prompt+answer sequence.
A linear projection head maps each $\mathbf{z}_i$ to logits over $P$ image patches:
\begin{equation}
\mathbf{g}_i = \mathbf{W}_g \mathbf{z}_i + \mathbf{b}_g \in \mathbb{R}^{P}.
\end{equation}

\paragraph{14-label classifier head.}
For multi-label prediction, we attach a linear classifier on the final hidden state (we use the last token representation $\mathbf{h}_{\text{last}}\in\mathbb{R}^{d}$):
\begin{equation}
\hat{\mathbf{y}} = \sigma(\mathbf{W}_c \mathbf{h}_{\text{last}} + \mathbf{b}_c)\in(0,1)^{14}.
\end{equation}

\subsection{Two-stage Training Objective}
We employ a two-stage optimization strategy.

\subsubsection{Stage 1: Gaze-supervised token learning}
Stage 1 focuses on learning a consistent mapping between the four gaze tokens and gaze-selected patch indices.
For each gaze token $i$, we are given a (possibly empty) list of target patch indices $\mathcal{P}_i=\{p_{i,1},\dots,p_{i,K_i}\}$.
We optimize a cross-entropy loss over patch IDs, masking tokens with missing gaze targets:
\begin{equation}
\mathcal{L}_{\text{gaze}} = \sum_{i=1}^{4}\ \mathbb{I}[K_i>0]\ \frac{1}{K_i}\sum_{j=1}^{K_i}\text{CE}(\mathbf{g}_i, p_{i,j}).
\end{equation}
Here, $\mathbf{g}_i$ denotes the predicted distribution over patch indices for token $i$.
The inner average normalizes the contribution of each token, so the loss magnitude does not scale with the number of gaze patches $K_i$.
When $K_i=0$, the corresponding term is dropped (i.e., no gaze supervision is applied to that token for the current sample).

In our implementation, we train only lightweight modules (LoRA~\cite{hu2022lora} adapters and the gaze projection head) to reduce memory footprint, while keeping the majority of the backbone frozen.

\subsubsection{Stage 2: Multi-label classification with joint language modeling}
Stage 2 trains a 14-label classifier head while maintaining the constrained answer format.
We compute a binary cross-entropy loss:
\begin{equation}
\mathcal{L}_{\text{cls}} = \text{BCE}(\hat{\mathbf{y}}, \mathbf{y}).
\end{equation}
Optionally, we combine $\mathcal{L}_{\text{cls}}$ with a language modeling loss $\mathcal{L}_{\text{lm}}$ from teacher-forced generation:
\begin{equation}
\mathcal{L} = (1-\lambda)\mathcal{L}_{\text{lm}} + \lambda \mathcal{L}_{\text{cls}}.
\end{equation}

\subsection{Implementation Details}
All experiments were conducted on a workstation with 8$\times$24GB NVIDIA RTX A6000 GPUs. To reduce memory usage, we use parameter-efficient fine-tuning with LoRA~\cite{devalal2018lora}, updating only lightweight adapters while keeping the pretrained backbone largely frozen. Images are processed by the VLM processor, with an optional maximum pixel budget to cap the number of visual tokens. Each training sample is formatted as a chat input with one image and an instruction prompt, and the assistant is supervised to produce a strict fixed-format answer string. We keep exactly four gaze placeholder tokens to consistently extract their hidden representations for gaze supervision. The loss weight in Eq.~(7) is set to $\lambda=0.7$.

\begin{figure}[!t]
  \centering
  \includegraphics[width=\linewidth]{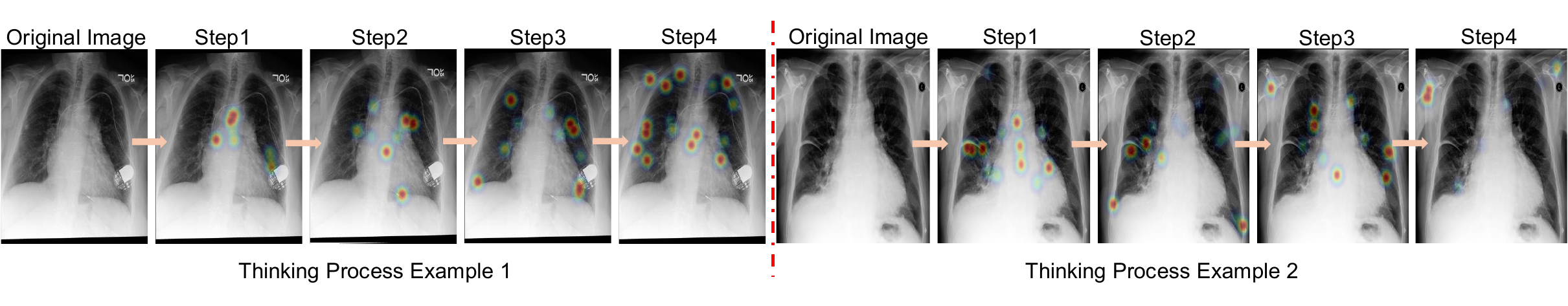}
  \caption{\textbf{Eye-gaze reasoning trajectories.} Two MIMIC-EYE examples showing temporally ordered gaze heatmaps overlaid on the chest X-ray. Each sequence visualizes how attention evolves from Step~1 to Step~4, illustrating radiologists' sequential evidence acquisition during interpretation.}
  \label{fig:exp1}
\end{figure}
\vspace{-0.2mm}
\begin{table}[t]
\centering
\caption{In-domain performance on MIMIC-EYE. Area under ROC curve (AUROC), Accuracy (Acc.) and F1-score (F1) are reported. Best results are in \textbf{bold} and second-best results are \underline{underlined}.}
\label{tab:mimic_eye_auroc_wide}
\setlength{\tabcolsep}{10pt}
\renewcommand{\arraystretch}{1.15}
\begin{tabular}{lccc}
\toprule
\textbf{Method} & \textbf{AUROC}$\uparrow$ & \textbf{Acc.}$\uparrow$ & \textbf{F1}$\uparrow$ \\
\midrule
Vanilla              & 49.74 & 42.15 & 43.09 \\
SFT                  & 87.60 & 86.03 & 84.18 \\
SFT-Heatmap          & 87.51 & 86.51 & 84.23 \\
MedCLIP~\cite{wang2022medclip}     & 87.37 & 86.63 & 84.32 \\
EGMA~\cite{ma2024eye}              & \underline{89.49} & \underline{88.11} & 86.20 \\
Random-Gaze          & 86.45 & 85.59 & 81.06 \\
Shuffled-Gaze        & 88.51 & 87.48 & \underline{84.97} \\
Original-Gaze        & \textbf{90.17} & \textbf{89.02} & \textbf{87.61} \\
\bottomrule
\end{tabular}
\end{table}
\vspace{-0.2mm}

\section{Experiments}

\subsection{In-domain Evaluation on MIMIC-EYE}
We evaluate our method on the MIMIC-EYE test set using AUROC (Table~\ref{tab:mimic_eye_auroc_wide}). 
Starting from the vanilla Qwen2.5VL-7B baseline, supervised fine-tuning (SFT) yields a large gain in AUROC (49.74 $\rightarrow$ 87.60), showing that task-specific instruction tuning is essential for reliable 14-label prediction. Visualization results can be seen in Figure~\ref{fig:exp1}.

Building on SFT, we incorporate eye-gaze supervision. A heatmap-based gaze signal provides a small but consistent improvement, suggesting that gaze offers complementary localization cues beyond text supervision.
More importantly, when gaze is injected as token-level patch supervision, preserving the original temporal order brings the strongest benefit.
Among all variants, \textbf{Original-Gaze} achieves the best in-domain performance, reaching 90.17 AUROC and outperforming both Shuffled-Gaze (88.51) and Random-Gaze (86.45).
This trend supports our hypothesis that gaze is not only a spatial prior: its sequential structure reflects expert evidence acquisition and aligns well with token-based VLM computation, leading to better calibrated and more accurate multi-label predictions.

\subsection{Zero-shot Generalization}
We further assess generalization by evaluating zero-shot classification on three external benchmarks (CheXpert 5$\times$200~\cite{irvin2019chexpert}, RSNA~\cite{rsna_pneumonia_ack_2018}, and SIIM-ACR~\cite{siim_acr_pneumothorax_2019}) using Accuracy and F1-score (Table~\ref{tab:zero_shot_comparison}). 
Across all datasets, SFT improves substantially over the vanilla model (e.g., CheXpert Acc 43.96 $\rightarrow$ 55.60; RSNA Acc 60.77 $\rightarrow$ 68.83), indicating strong transfer from MIMIC-EYE instruction tuning. Incorporating gaze supervision yields additional consistent gains.

Our \textbf{Original-Gaze} is the best-performing method on every benchmark, achieving 62.45/61.73 (Acc/F1) on CheXpert 5$\times$200, 77.61/53.73 on RSNA, and 64.07/61.89 on SIIM-ACR.
Ablations further highlight the role of structured gaze: both Random-Gaze and Shuffled-Gaze often outperform SFT, but preserving the original gaze order yields the most consistent and largest improvements, especially on the harder metrics such as F1 (e.g., RSNA F1 48.64 $\rightarrow$ 53.73).
Overall, these results demonstrate that gaze-supervised token learning improves not only in-domain accuracy but also out-of-domain robustness, suggesting that ordered gaze encodes transferable visual evidence cues that benefit VLM-based medical classifiers.
\vspace{-0.2mm}

\begin{table}[t]
\centering
\caption{Comparison results of zero-shot classification tasks on CheXpert 5$\times$200~\cite{irvin2019chexpert}, RSNA~\cite{rsna_pneumonia_ack_2018}, and SIIM-ACR~\cite{siim_acr_pneumothorax_2019} datasets. Accuracy (Acc.) and F1-score (F1) are reported. Vanilla: Qwen2.5VL-7B. SFT: Qwen2.5VL-7B-SFT. SFT-Heatmap: Qwen2.5VL-7B-SFT with Eye-Gazed heatmap. Random-Gaze: replace gaze patch IDs with random indices, preserving list sizes. Shuffled-Gaze: permute the gaze lists. Best results are in \textbf{bold} and second-best results are \underline{underlined}.}
\label{tab:zero_shot_comparison}
\setlength{\tabcolsep}{6pt}
\begin{tabular}{lcc cc cc}
\toprule
\multirow{2}{*}{Method} &
\multicolumn{2}{c}{CheXpert 5$\times$200} &
\multicolumn{2}{c}{RSNA} &
\multicolumn{2}{c}{SIIM-ACR} \\
\cmidrule(lr){2-3}\cmidrule(lr){4-5}\cmidrule(lr){6-7}
& Acc.$\uparrow$ & F1$\uparrow$ & Acc.$\uparrow$ & F1$\uparrow$ & Acc.$\uparrow$ & F1$\uparrow$ \\
\midrule
Vanilla     & 43.96 & 41.56 & 60.77 & 37.80 & 50.12 & 37.66 \\
SFT         & 55.62 & 51.37 & 68.83 & 43.77 & 60.03 & 55.45 \\
SFT-Heatmap & 58.39 & 55.99 & 71.10 & 44.73 & 58.40 & 57.85 \\
MedCLIP~\cite{wang2022medclip} & 57.50 & 55.97 & 73.09 & 41.01 & 58.76 & 59.16 \\
EGMA~\cite{ma2024eye}          & 59.30 & \underline{60.38} & \textbf{76.97} & 46.12 & \underline{63.62} & \underline{61.46} \\
Random-Gaze                      & 56.17 & 53.92 & 69.06 & 43.39 & 60.31 & 57.72 \\
Shuffled-Gaze                    & \underline{59.33} & 57.38 & 73.97 & \underline{48.64} & 61.62 & 58.46 \\
Original-Gaze                    & \textbf{62.45} & \textbf{61.73} & \underline{76.61} & \textbf{53.73} & \textbf{64.07} & \textbf{61.89} \\
\bottomrule
\end{tabular}
\end{table}

\section{Conclusion}
In this work, we make VLM-based chest X-ray interpretation more visually grounded using temporally ordered eye-gaze supervision. Under a fixed-format reporting interface, dedicated gaze tokens are trained to predict gaze-selected patch indices in order, encouraging radiologist-like evidence acquisition.

In-domain and zero-shot evaluations show consistent gains from preserving gaze order, indicating that eye-gaze provides transferable temporal cues that improve accuracy and robustness under distribution shift. The resulting gaze-linked evidence also supports clinician-friendly auditing.

\bibliographystyle{splncs04}
\bibliography{mybibliography}






\end{document}